\documentclass[final]{cvpr}
\usepackage{nopageno}

\usepackage{times}
\usepackage{epsfig}
\usepackage{graphicx}
\usepackage{amsmath}
\usepackage{amssymb}
\usepackage{booktabs}
\usepackage{multirow}

\DeclareMathOperator{\diag}{diag}

\usepackage[pagebackref=true,breaklinks=true,colorlinks,bookmarks=false]{hyperref}

\def\cvprPaperID{****} 
\def\confYear{CVPR 2021}
\begin{document}

\title{RAFT-3D: Scene Flow using Rigid-Motion Embeddings}

\author{Zachary Teed \ \ and \ \ Jia Deng \\
Princeton University \\
{\tt\small \{zteed,jiadeng\}@cs.princeton.edu}}

\maketitle

\begin{abstract}
  We address the problem of scene flow: given a pair of stereo or RGB-D video frames, estimate pixelwise 3D motion. We introduce RAFT-3D, a new deep architecture for scene flow. RAFT-3D is based on the RAFT model developed for optical flow but iteratively updates a dense field of pixelwise SE3 motion instead of 2D motion. A key innovation of RAFT-3D is rigid-motion embeddings, which represent a soft grouping of pixels into rigid objects. Integral to rigid-motion embeddings is Dense-SE3, a differentiable layer that enforces geometric consistency of the embeddings. Experiments show that RAFT-3D achieves state-of-the-art performance. On FlyingThings3D, under the two-view evaluation, we improved the best published accuracy ($\delta < 0.05$) from 34.3\% to 83.7\%. On KITTI, we achieve an error of 5.77, outperforming the best published method (6.31), despite using no object instance supervision. Code is available at \url{https://github.com/princeton-vl/RAFT-3D}.
  \end{abstract}

\section{Introduction}

\begin{figure*}
    \centering
    \includegraphics[width=.9\linewidth]{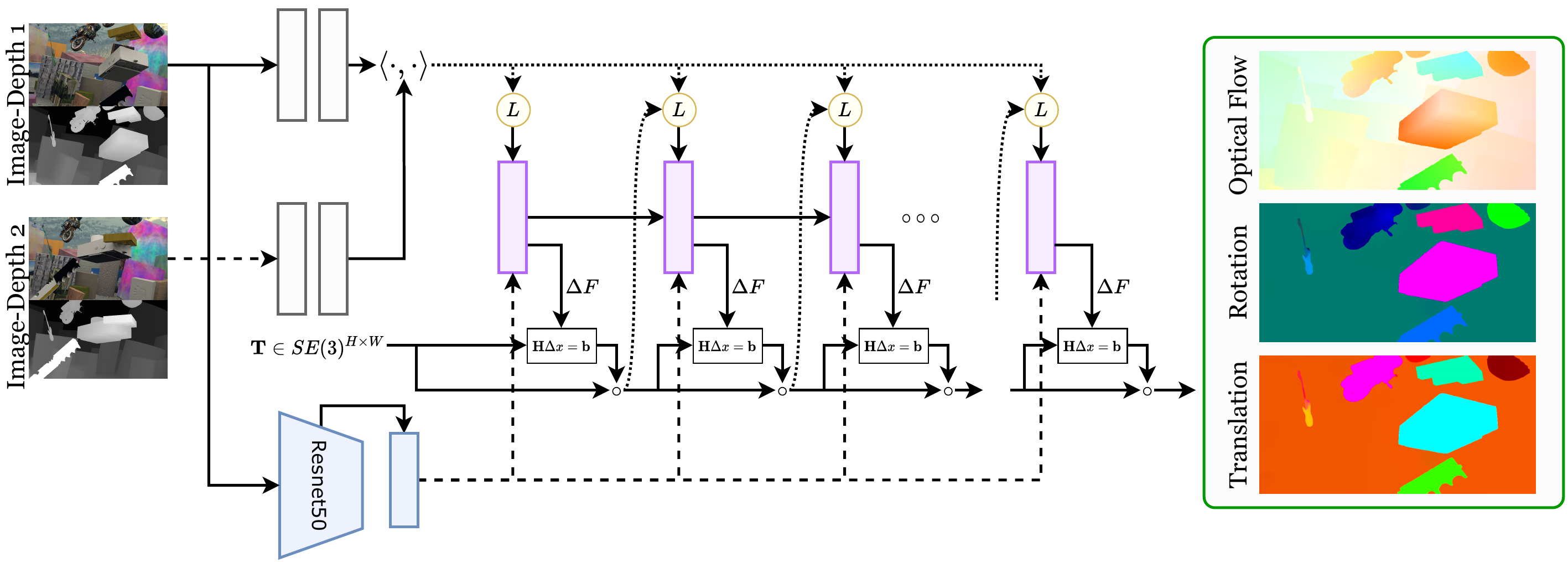}
    \caption{Overview of our approach. Features extracted from the input images are used to construct a 4D correlation volume. We initialize the SE3 motion field, $\mathbf{T}$, to be the identity at every pixel. During each iteration, the update operator uses the current SE3 motion estimate to index from the correlation volume, using the correlation features and hidden state to produce estimates of pixel correspondence and rigid-motion embeddings. These estimates are plugged into Dense-SE3, a least-squares optimization layer which uses geometric constraints to produce an update to the SE3 field. After successive iterations we recover a dense SE3 field, which can be decomposed into a rotational and translation component. The SE3 field can be projected onto the image to recover optical flow.}
    \label{fig:RAFT3D}
\end{figure*}

Scene flow is the task of estimating pixelwise 3D motion between a pair of video frames\cite{vedula1999three}. Detailed 3D motion is requisite for many downstream applications including path planning, collision avoidance, virtual reality, and motion modeling. In this paper, we focus on stereo scene flow and RGB-D scene flow, which address stereo video and RGB-D video respectively. 

Many scenes can be well approximated as a collection of rigidly moving objects. The motion of driving scenes, for example, can be modeled as a variable number of cars, buses, and trucks. The most successful scene flow approaches have exploited this structure by decomposing a scene into its rigidly moving components\cite{menze2015object,vogel20113d,vogel20113d,vogel20153d,ma2019deep,behl2017bounding,jaimez2015motion,jaimez2015primal,kumar2017monocular}. This introduces a powerful prior which can be used to guide inference. While optical flow approaches typically assume piecewise smooth motion, a scene containing rigid objects will exhibit piecewise constant 3D motion fields (Fig. \ref{fig:RAFT3D}).

Recently, many works have proposed integrating deep learning into scene flow estimation pipelines. A common approach has been to use object detection\cite{behl2017bounding,cao2019learning} or segmentation \cite{behl2017bounding,ma2019deep,lv2018learning,ren2017cascaded} networks to decompose the scene into a collection of potentially rigidly moving objects. Once the scene has been segmented into its rigidly moving components, more traditional optimization can be used to fit a motion model to each of the objects. One limitation of this approach is that the networks require instance segmentations to be trained and cannot recover the motion of new unknown objects. Object detection and instance segmentation introduce non-differentiable components into the network, making end-to-end training difficult without bounding box or instance level supervision.

We introduce RAFT-3D, an end-to-end differentiable architecture which estimates pixelwise 3D motion from stereo or RGB-D video. RAFT-3D is built on top of RAFT~\cite{teed2020raft}, a state-of-the-art optical flow architecture that builds all-pairs correlation volumes and uses a recurrent unit to iteratively refine a 2D flow field. We retain the basic iterative structure of RAFT but introduce a number of novel designs.

The main innovation we introduce is rigid-motion embeddings, which are per-pixel vectors that represent a soft grouping of pixels into rigid objects.  During inference, RAFT-3D iteratively updates the rigid-motion embeddings such that pixels with similar embeddings belong to the same rigid object and follow the same SE3 motion. 

Integral to rigid-motion embeddings is \emph{Dense-SE3}, a differentiable layer that seeks to ensure that the embeddings are geometrically meaningful. Dense-SE3 iteratively updates a dense field of per-pixel SE3 motion by performing unrolled Gauss-Newton iterations such that the per-pixel SE3 motion is geometrically consistent with the current estimates of rigid-motion embeddings and pixel correspondence. Because of Dense-SE3,  the rigid-motion embeddings can be indirectly supervised from only ground truth 3D scene flow, and our approach does not need any supervision of object boxes or masks. 

Fig. \ref{fig:RAFT3D} provides an overview of our approach. RAFT-3D take a pair of RGB-D images as input. It extracts features from the input images and builds a 4D correlation volume by computing the visual similarity between all pairs of pixels.  RAFT-3D maintains and updates a dense field of pixelwise SE3 motion. During each iteration, it uses the current estimate of SE3 motion to index from the correlation volume. A recurrent GRU-based update operator takes the correlation features and produces an estimate of pixel correspondence, which is then used by Dense-SE3 to generate updates to the SE3 motion field. 

RAFT-3D achieves state-of-the-art accuracy. 
On FlyingThings3D, under the two-view evaluation~\cite{liu2019flownet3d}, RAFT-3D improves the best published accuracy ($\delta < 0.05$) from 34.3\% to 83.7\%. On KITTI, RAFT-3D achieves an error of 5.77, outperforming the best published method (6.31), despite using no object instance supervision. 

\section{Related Work}

The task of reconstructing a 3D motion field from video is often referred to as estimating ``scene flow''. 

\vspace{1mm} \noindent \textbf{Optical Flow:} Optical flow is the problem of estimating dense 2D pixel-level motion between a pair of frames. Early work formulated optical flow as a energy minimization problem, where the objective was a combination of a data term---encouraging the matching of visually similar image regions---and a regularization term---favoring piecewise smooth motion fields. Many early scene flow approaches evolved from this formulation, replacing piecewise \emph{smooth} flow priors with a piecewise \emph{constant} rotation/translation field prior\cite{vogel2013piecewise,menze2015object}. This greater degree of structure allowed scene flow methods to outperform approaches which treated optical flow or stereo separately\cite{vogel20113d}.

Recently, the problem of optical flow has been reformulated in the context of deep learning. Many works have demonstrated that a neural network can be directly trained to estimate optical flow between a pair of frames, and a large variety of network architectures have been proposed for the task \cite{flownet1,flownet2,pwcnet,ranjan2017optical,lu2020devon,yang2019volumetric,teed2020raft}. RAFT\cite{teed2020raft} is a recurrent network architecture for estimating optical flow. RAFT builds a 4D correlation volume by computing the visual similarity between all pairs of pixels; then, during inference, a recurrent update operator indexes from the correlation volume to produce a flow update. A unique feature of RAFT is that a single, high resolution, flow field is updated and maintained. 

Our approach is based on the RAFT architecture, but instead of a flow field, we estimate a SE3 motion field, where a rigid body transformation is estimated for each pixel. When projected onto the image, our SE3 motion vectors give more accurate optical flow than RAFT.

\vspace{1mm} \noindent \textbf{Rectified Stereo:} Rectified stereo can be viewed as a 1-dimensional analog to optical flow, where the correspondence of each pixel in the left image is constrained to lie on a horizontal line spanning the right image. Like optical flow, traditional methods treated stereo as an energy minimization problem\cite{hirschmuller2005accurate,ranftl2012pushing} often exploiting planar information\cite{bleyer2011patchmatch}. 

Recent deep learning approaches have borrowed many core concepts from conventional approaches such as the use of a 3D cost volume \cite{gcnet}, replacing hand-crafted features and similarity metrics with learned features, and cost volume filtering with a learned 3D CNN. Like optical flow, a variety of network architectures have been proposed \cite{gcnet,zhang2019ga,guo2019group,chang2018pyramid}. Here we use GA-Net\cite{zhang2019ga} to estimate depth between the each left/right image pair.

\vspace{1mm} \noindent \textbf{Scene Flow:} Like optical flow and stereo, scene flow can be approached as a energy minimization problem. The objective is to recover a flow field such that (1) visually similar image regions are aligned and (2) the flow field maximizes some prior such as piecewise rigid motion and piecewise planar depth. Both variational optimization\cite{quiroga2014dense,jaimez2015motion,jaimez2015primal} and discrete optimization\cite{menze2015object,jaimez2015primal} approaches have been explored for inference. Our network is designed to mimic the behavior an optimization algorithm. We maintain an estimate of the current motion field which is updated and refined with each iteration.

Jaimez et al.\cite{jaimez2015motion} proposed an alternating optimization approach for scene flow estimation from a pair of RGB-D images, iterating between grouping pixels into rigidly moving clusters and estimating the motion model for each of the cluster. Our method shares key ideas with this approach, namely the grouping of pixels into rigidly moving objects, however, we avoid a hard clustering by using rigid-motion embeddings, which softly and differentiably group pixels into rigid objects.

Recent works have leveraged the object detection and semantic segmentation ability of deep networks to improve scene flow accuracy\cite{ma2019deep,cao2019learning,ren2017cascaded,behl2017bounding,gordon2019depth}. In these works, an object detection or instance segmentation network is trained to identify potentially moving objects, such as cars or buses. While these approaches have been very effective for driving datasets such as KITTI where moving objects can be easily identified using semantics, they do not generalize well to novel objects. An additional limitation is that the detection and instance segmentation introduces non-differentiable components into the pipeline, requiring these components to be trained separately on ground truth annotation. Ma et al. \cite{ma2019deep} was able to train an instance segmentation network jointly with optical flow estimation by differentiating through Gauss-Newton updates; however, this required additional instance supervision and pre-training on Cityscapes\cite{cordts2016cityscapes}. On the other hand, our network outperforms these approaches without using object instance supervision.

Yang and Ramanan\cite{yang2020upgrading} take a unique approach and use a network to predict optical expansion, or the change in perceived object size. Combining optical expansion with optical flow gives normalized 3D scene flow. The scale ambiguity can be recovered using Lidar, stereo, or monocular depth estimation. This approach does not require instance segmentation, but also cannot directly enforce rigid motion priors.

Another line of work has focused on estimating 3D motion between a pair \cite{liu2019flownet3d,wang2020flownet3d++,gu2019hplflownet} or sequence\cite{liu2019meteornet,fan2019pointrnn} of point clouds. These approaches are well suited for Lidar data where the sensor produces sparse measurements. However, these works do not directly exploit scene rigidity. As we demonstrate in our experiments, reasoning about object level rigidity is critical for good accuracy.

\section{Approach}

We propose an iterative architecture for scene flow estimation from a pair of RGB-D images. Our network takes in two image/depth pairs, $(I_1, Z_1)$, $(I_2, Z_2)$, and outputs a dense transformation field $\mathbf{T} \in SE(3)^{H \times W}$ which assigns a rigid body transformation to each pixel. For stereo images, the depth estimates $Z_1$ and $Z_2$ are obtained using an off-the-shelf stereo network.

\subsection{Preliminaries}
We use the pinhole projection model and assume known camera intrinsics. We use an augmented projection function which maps a 3D point to its projected pixel coordinates, $(x,y)$, in addition to inverse depth $d = 1/Z$. Given a homogeneous 3D point $\mathbf{X} = (X, Y, Z, 1)$
\begin{equation}
    (x, y, d) = \pi(\mathbf{X}) = \begin{pmatrix} f_x (X/Z) + c_x \\ f_y (Y/Z) + c_y \\ 1/Z \end{pmatrix}
\end{equation}
where $(f_x, f_y, c_x, c_y)$ are the camera intrinsics.

Given a dense depth map $Z \in \mathbb{R_+}^{H\times W}$, we can use the inverse projection function.
\begin{equation}
    \begin{pmatrix} X \\ Y \\ Z \\ 1 \end{pmatrix} = \pi^{-1}(x,y,d) = \frac{1}{d} \begin{pmatrix} (x-c_x)/f_x \\ (x-c_y)/f_y \\ 1 \\ d  \end{pmatrix}
\end{equation}
which maps from pixel $(x, y, d)$ to the point $(X, Y, Z, 1)$, again with inverse depth $d = 1/z$.

\vspace{1mm} \noindent \textbf{Mapping Between Images:} We use a dense transformation field, $\mathbf{T} \in SE(3)^{H \times W}$ to represent the 3D motion between a pair of frames. Using $\mathbf{T}$, we can construct a function which maps points in frame $I_1$ to $I_2$. Letting $\mathbf{x}_{i}=(x_{i},y_{i},d_{i})$ be the pixel coordinate at index $i$ then the mapping
\begin{equation}
    \mathbf{x}'_{i} = (x_i', y_i', d_i') = \pi(\mathbf{T}_{i} \cdot \mathbf{X}_{i}), \qquad \mathbf{X}_{i} = \pi^{-1}(\mathbf{x}_{i})
    \label{eqn:mapping}
\end{equation}
can be used to find the correspondence of $\mathbf{x}_{i}$ in $I_2$.

A flow vector can be obtained by taking the difference $\mathbf{x}'_{i} - \mathbf{x}_{i}$. The first two components of the flow vector give us the standard optical flow. The last component provides the change in inverse depth between the pair of frames. The focus of this paper is to recover $\mathbf{T}$ given a pair of frames.

\vspace{1mm} \noindent \textbf{Jacobians:} For optimization purposes, we will need the Jacobian of the Eqn. \ref{eqn:mapping}. Using the chain rule, we can compute the Jacobian of Eqn. \ref{eqn:mapping} as the product of the projection Jacobian
\begin{equation}
    \mathbf{J}_{\pi} = \frac{\partial \pi(\mathbf{X}')}{\partial \mathbf{X}'} =
    \begin{pmatrix} 
    f_x d' & 0 & -f_x X' {d'}^2 \\ 
    0 & f_y d' & -f_y Y' {d'}^2 \\
    0 & 0 & -{d'}^2\end{pmatrix}
\end{equation}
and the transformation Jacobian
\begin{equation}
    \mathbf{J}_{T} = 
    \left(\mathbf{I}_{3\times 3},  (\mathbf{X}')^\wedge \right), \ \ \ \mathbf{w}^\wedge = \begin{pmatrix} 0 & \text{-}w_3 & w_2 \\ w_3 & 0 & \text{-}w_1 \\ \text{-}w_2 & w_1 & 0 \end{pmatrix}
\end{equation}
using local coordinates defined by the retraction $\exp(\boldsymbol{\delta}^\wedge)\cdot\mathbf{T}$. Giving the Jacobian of Eqn. \ref{eqn:mapping} as $\mathbf{J} = \mathbf{J}_\pi \cdot \mathbf{J}_T \in \mathbb{R}^{3\times 6}$.

\vspace{1mm} \noindent \textbf{Optimization on Lie Manifolds:} The space of rigid-body transformations forms a Lie group, which is a smooth manifold and a group. In this paper, we use the Gauss-Newton algorithm to perform optimization steps over the space of dense SE3 fields. 

Given a weighted least squares objective
\begin{equation}
    E(\mathbf{x}) = \sum_i w_i \cdot (f_i(\mathbf{x}) - y_i)^2
\end{equation}
the Gauss-Newton algorithm linearizes the residual terms, and solves for the update
\begin{align}
    &\mathbf{J}^T \ \diag(\mathbf{w}) \ \mathbf{J} \Delta \mathbf{x} = \ \mathbf{J}^T \mathbf{r}(\mathbf{x}) \\
    &r_i = f_i(\mathbf{x}) - y_i \qquad \mathbf{J}_i = \left. \frac{\partial f_i(\exp(\delta^\wedge) \mathbf{x})}{\partial \delta} \right|_{\delta=0} 
    \label{eqn:update}
\end{align}
The update is found by solving Eqn.~\ref{eqn:update} and applying the retraction $\mathbf{T}' = \exp(\Delta \mathbf{x}^\wedge) \cdot \mathbf{T}$. Eqn.~\ref{eqn:update} can be rewritten as the linear system
\begin{equation}
    \mathbf{H} \Delta \mathbf{x} = \mathbf{b} \qquad 
    \mathbf{H} = \mathbf{J}^T \ \diag(\mathbf{w}) \ \mathbf{J}, \ 
    \mathbf{b} = \mathbf{J}^T \mathbf{r}(\mathbf{x})
\end{equation}
and $\mathbf{H}$ and $\mathbf{b}$ can be constructed without explicitly forming the Jacobian matrices
\begin{equation}
    \mathbf{H} = \sum_i w_i \cdot \mathbf{J}_i^T \mathbf{J}_i, \qquad \mathbf{b} = \sum_i w_i \cdot  \mathbf{J}_i^T r_i(\mathbf{x}).
    \label{eqn:inplace}
\end{equation}
This fact is especially useful when solving optimization functions with millions of residual terms. In this setting, storing the full Jacobian matrix becomes impractical.

\subsection{Network Architecture}
Our network architecture is based on RAFT\cite{teed2020raft}. We construct a full 4D correlation volume by computing the visual similarity between all pairs of pixels between the two input images. During each iteration, the network uses the current estimate of the SE3 field to index from the correlation volume. Correlation features are then fed into an recurrent update operator which estimates a dense flow field. We provide an overview of the RAFT architecture here, but more details can be found in \cite{teed2020raft}.

\vspace{1mm} \noindent \textbf{Feature Extraction:}
We first extract features from the two input images. We use two separate feature extract networks. The feature encoder, $f_\theta$, is applied to both images with shared weights. $f_\theta$ extracts a dense 128-dimension feature vector at 1/8 resolution. It consists of 6 residuals blocks, 2 at 1/2 resolution, 2 at 1/4 resolution, and 2 at 1/8 resolution. We provide more details of the network architectures in the appendix.

The context encoder extracts semantic and contextual information from the first image. Different from the original RAFT\cite{teed2020raft}, we use a pretrained ResNet50\cite{resnet} with a skip connection to extract context features at 1/8 resolution. The reason behind this change is that grouping objects into rigidly moving regions requires a greater degree of semantic information and larger receptive field. During training, we freeze the batch norm layers in the context encoder.

\vspace{1mm} \noindent \textbf{Computing Visual Similarity:} We construct a 4D correlation volume by computing the dot product between all-pairs of feature vectors between the input images
\begin{equation}
    \mathbf{C}_{ijkh}(I_1, I_2) = \langle f_\theta(I_1)_{ij}, \ f_\theta(I_2)_{kh} \rangle \in \mathbb{R}^{H \times W \times H \times W}
\end{equation}
We then pool the last two dimensions of the correlation volume 3 times using average pooling with a $2\times 2$ kernel, resulting in a correlation pyramid $\{\mathbf{C}_1, \mathbf{C}_2, \mathbf{C}_3, \mathbf{C}_4\}$ with
\begin{equation}
    \mathbf{C}_k \in \mathbb{R}^{H \times W \times H/2^{k-1} \times W/2^{k-1}}
\end{equation}

\vspace{1mm} \noindent \textbf{Indexing the Correlation Pyramid: } Given a current estimate of correspondence $\mathbf{x}'=(u,v)$, we can index from the correlation volume to produce a set of correlation features. First we construct a neighborhood grid around $\mathbf{x}$
\begin{equation}
    \mathcal{N}_{\mathbf{x}} = \{(u + d_u, v+d_v) \ | \ d_u, d_v \in \{-r, ..., r\} \ \} \ \
\end{equation}
and then use the neighboorhood to sample from the correlation volume using bilinear sampling. We note that the constructing and indexing from the correlation volume is performed in an identical manner to RAFT\cite{teed2020deepv2d}.

\vspace{1mm} \noindent \textbf{Update Operator:} The update operator is a recurrent GRU-unit which retrieves features from the correlation volume using the indexing operator and outputs a set of revisions. RAFT uses a series of 1x5 and 5x1 GRU units; we use a single 3x3 unit but use a kernel composed of 1 and 3 dilation rates. We provide more details on the architecture of the update operator in the appendix.

Using Eqn. \ref{eqn:mapping}, we can use the current estimate of $\mathbf{T}$ to estimate 2D correspondences $\mathbf{x}' = \pi(\mathbf{T} \cdot \pi^{-1}(\mathbf{x}))$. The following features are used as input to the GRU
\begin{itemize}
\item[--] Flow field:  $\mathbf{x}' - \mathbf{x}$
\item[--] Twist field:  $\log_{SE3} (\mathbf{T})$
\item[--] Depth residual: $\mathbf{d}' - \mathbf{\bar{d}}'$
\item[--] Correlation features: $L_\mathbf{C}(\mathbf{x}')$
\end{itemize}
In the depth residual term, the inverse depth $d'_i$ is obtained from the depth component of $\mathbf{x}_i'$, i.e.\@ the backprojected pixel $i$ expressed in the coordinate system of frame 2. The inverse depth $\bar{d}'_i$ is obtained by indexing the inverse depth map of frame 2 using
the correspondence $x_i'$ of pixel $i$.
If pixel $i$ is non-occluded, an accurate SE3 field $\mathbf{T}$ should result in a depth residual of 0. Each of the derived features are processed through 2 convolutional layers and then provided as input to the convolutional GRU.

The hidden state is then used to predict the inputs to the Dense-SE3 layer. We apply two convolutional layers to hidden state to output a rigid-motion embedding map $\mathbf{V}$. We additionally predict a ``revision map'' $\mathbf{r}_x, \mathbf{r}_y, \mathbf{r}_z$ and corresponding confidence maps $\mathbf{w}_x, \mathbf{w}_y, \mathbf{w}_z \in [0,1]$. The revisions $\mathbf{r}_x$ and $\mathbf{r}_y$ correspond to corrections that should be   made to the optical flow induced by the current SE3 field. In other words, the network is trying to get a new estimate of pixel correspondence, but is expressing it on top of the flow induced by the SE3 field. 
The revisions $\mathbf{r}_z$ is the corrections that should be made to the inverse depth in frame 2 when the inverse depth is used by Dense-SE3 to enforce geometric consistency. This is to account for noise in the input depth as well as occlusions. The embedding map and revision maps are taken as input to the Dense-SE3 layer to produce an update to the SE3 motion field.

\vspace{1mm} \noindent \textbf{SE3 Upsampling:} The SE3 motion field estimated by the network is at 1/8 of the resolution. We use convex up-sampling \cite{teed2020raft} to upsample the transformation field to the full input resolution. In RAFT\cite{teed2020raft}, the high resolution flow field was taken to be the convex combination of $3 \times 3$ grids at the lower resolution with combination weights predicted by the network. However, the SE3 field $\mathbf{T}$ lies on a manifold and is not closed under linear combinations. Instead we perform upsampling by first mapping $\mathbf{T}$ to the Lie algebra using the logarithm map, performing convex upsampling in the lie algebra, and then mapping back to the manifold using the exponential map.

\subsection{Dense-SE3 Layer}
The key ingredient to our approach is the Dense-SE3 layer. Each application of the update module produces a revision map $\mathbf{r}=(\mathbf{r}_x, \mathbf{r}_y, \mathbf{r}_z)$. The Dense-SE3 layer is a differentiable optimization layer which maps the revision map to a SE3 field update.

The rigid-motion embedding vectors are used to softly group pixels into rigid objects. Given two embedding vectors $\mathbf{v}_i$ and $\mathbf{v}_j$, we compute an affinity $a_{ij} \in [0,1]$ by taking the sigmoid of the negative L2 distance
\begin{equation}
    a_{ij} = 2 * \sigma(-||\mathbf{v}_i - \mathbf{v}_j||^2) \in [0,1]
\end{equation}

\noindent \textbf{Objective Function:} Using the affinity terms, we define an objective function based on the reprojection error
\begin{align}
    E(\delta) &= \sum_{i \in \Omega} \sum_{j \in \mathcal{N}_i} a_{ij} e^2_{ij}(\delta_i) \\
    e^2_{ij}(\delta_i) &= ||\mathbf{r}_j + \pi(\mathbf{T}_j \mathbf{X}_j) - \pi(e^{\delta_i} \mathbf{T}_i \mathbf{X}_j)||^2_{w_j}
    \label{eqn:objective}
\end{align}
with $||\mathbf{x}||_\mathbf{w}^2 = \mathbf{x}^T \diag(\mathbf{w}) \mathbf{x}$. The objective states that for every pixel $i$, we want a transformation $\mathbf{T}_i$ which describes the motion of pixels in a neighborhood $j \in \mathcal{N}_i$. However, not every pixel $j \in \mathcal{N}_i$ belongs to the same rigidly moving object. That is the purpose for the embedding vector. Only pairs $(i,j)$ with similar embeddings significantly contribute to the objective function. 

\vspace{1mm} \noindent \textbf{Efficient Optimization:} We apply a single Gauss-Newton update to Eqn. \ref{eqn:objective} to generate the next SE3 estimate. Since the Dense-SE3 layer is applied after each application of the update operator, 12 iterations of the update operator yields 12 Gauss-Newton updates.

The objective defined in Eqn. \ref{eqn:objective} can result in a very large optimization problem. We generally use a large neighborhood $\mathcal{N}_i$ in practice; in some experiments we take $\mathcal{N}_i$ to be the entire image. For the FlyingThings3D dataset, with $540\times960$ resolution, this results in \emph{200 million} equations and 50,000 variables (Dense-SE3 layer operators at 1/8 the input resolution). Trying the store the full system would exceed available memory. 

However, each term in Eqn. \ref{eqn:objective} only includes a single $\mathbf{T}_i$. This means that instead of solving a single optimization problem with $H \times W \times 6$ variables, we can instead solve a set of $H \times W$ problems each with only $6$ variables. Furthermore, we can leverage Eqn. \ref{eqn:inplace} and build the linear system in place without explicitly constructing the Jacobian. When implemented directly in Cuda, a Gauss-Newton update of Eqn. \ref{eqn:objective} can be performed very quickly and is not a bottleneck in our approach.

\begin{figure*}
    \small \hspace{-8mm} image \hspace{28mm} flow \hspace{30mm} $\tau$ \hspace{30mm} $\phi$
    \centering
    \includegraphics[width=.8\linewidth]{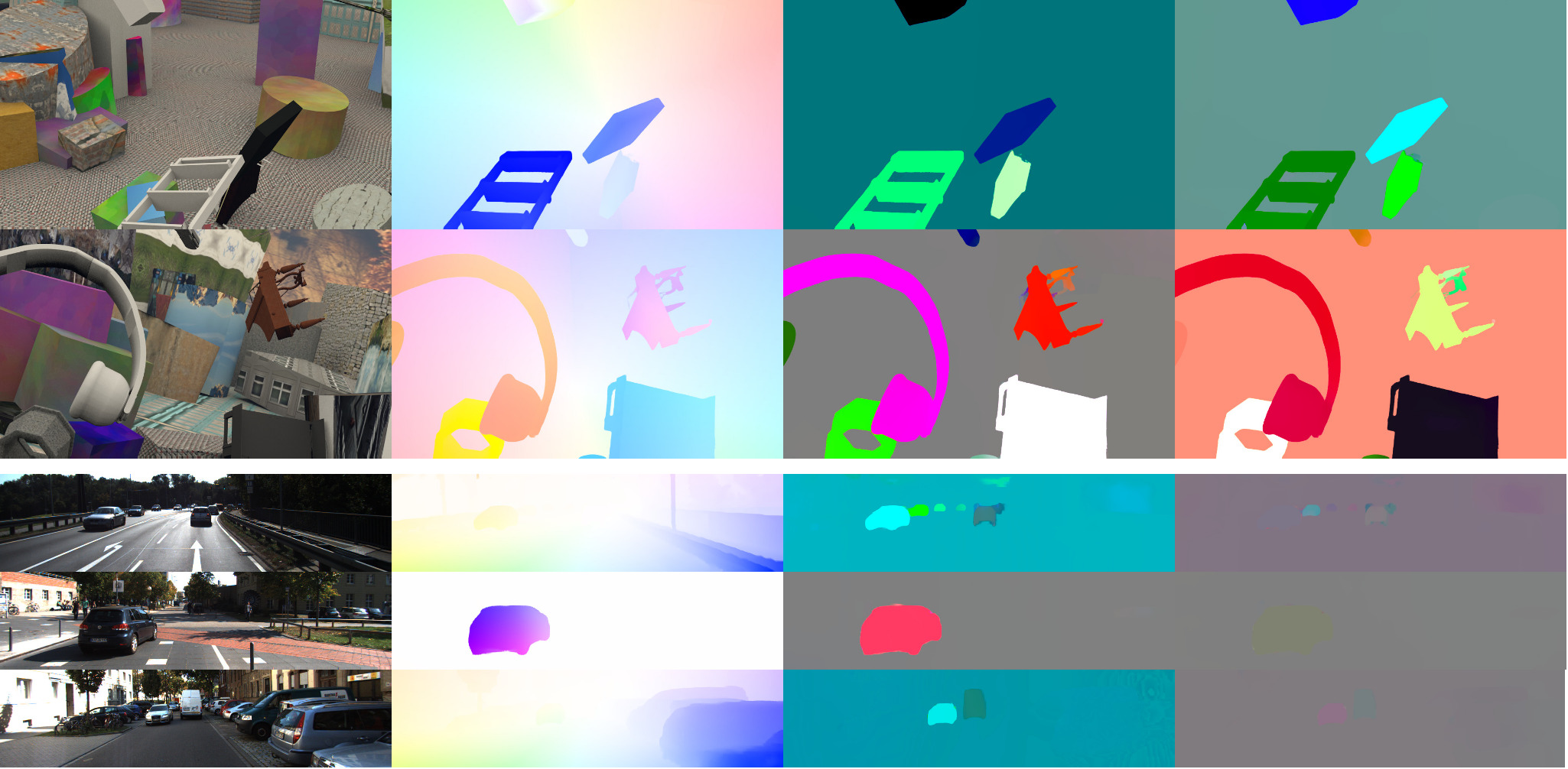}
    \caption{Visualization of the predicted motion fields on FlyingThings3D (top) and KITTI (bottom). Our network outputs a dense SE3 motion field,  which can be used to compute optical flow. We visualize the SE3 field as the twist field where $(\tau, \phi) = \log_{SE3}(\mathbf{T})$. Note that the twist fields are piecewise constant---pixels from the same rigid object are assigned the same SE3 motion.}
    \label{fig:examples}
\end{figure*}

\subsection{bi-Laplacian Embedding Optimization}
Since our architecture operates primarily at high resolution, it can be difficult for the network to group pixels which span large objects. We implement a differentiable bi-Laplacian optimization layer in order to smooth embedding vectors within motion boundaries. Vogel et al. \cite{vogel2018learning} used a similar differentiable optimization layer to smooth optical flow within motion boundaries; however, they use iterative methods to solve the linear system while we use direct Cholesky factorization which allows us to reuse the factorization for each channel of the embedding vector.

Given an embedding map $\mathbf{V} \in \mathbb{R}^{H \times W \times C}$, we have the GRU predict additional edge weights $\mathbf{w}_x, \mathbf{w}_y \in \mathbb{R}_+^{H \times W}$ and define the objective
\begin{equation}
    \mathbf{u}^* = \min_\mathbf{u} \left\{ ||D_x \mathbf{u}||_{\mathbf{w}_x}^2 + ||D_x \mathbf{u}||_{\mathbf{w}_y}^2 + ||\mathbf{u} - \mathbf{v}||^2 \right\}
    \label{eqn:bilaplacian}
\end{equation}
where $D_x$ and $D_y$ are linear finite difference operators, and $\mathbf{v}$ is the flattened feature map. 

In other words, we want to solve for a new embedding map $\mathbf{u}$ which is smooth within motion boundaries and close to the original embedding map $\mathbf{v}$. At boundaries, the network can set the weights to 0 so that edges do not get smoothed over. Eqn.~\ref{eqn:bilaplacian} can be solved in closed form using sparse Cholesky decomposition and we use the Cholmod library\cite{chen2008algorithm}. Using nested dissection\cite{george1973nested} factorization can be performed in $O({(HW)}^{1.5})$ time and backsubstition can be performed in $O({C\cdot(HW)}^{1.5})$ time. In the appendix, we derive the gradients of Eqn.~\ref{eqn:bilaplacian}. Since the optimization layer is differentiable, the inputs $\mathbf{w}_x$ and $\mathbf{w}_y$ don't require direct supervision.

\subsection{Supervision}
We supervise our network on a combination of ground truth optical flow and inverse depth change. Our network outputs a sequences of $\{\mathbf{T}_1, \mathbf{T}_2, \hdots,  \mathbf{T}_K \}$. For each transformation, $\mathbf{T}_k$, we computed the induced optical flow and inverse depth change
\begin{equation}
    \mathbf{f}_{est}^k = \pi(\mathbf{T}_k \cdot \pi^{-1}(\mathbf{x})) - \mathbf{x}
\end{equation}
where $\mathbf{x}$ is a dense coordinate grid in $I_1$. We compute the loss as the sequence over all estimations
\begin{equation}
    \mathcal{L} = \sum_{k=1}^N \gamma^{N-k}||\mathbf{f}_{est}^k - \mathbf{f}_{gt} {||}_1
\end{equation}
with $\gamma=0.9$. Note that no supervision is applied to the embedding vectors, and that rigid-motion embeddings are implicitly learned by differentiating through the dense $SE(3)$ update layer. We also apply an additional loss directly to the revisions predicted by the GRU with 0.2 weight.

\begin{table*}
\centering
\resizebox{.72\textwidth}{!}{
\begin{tabular}{lcccccc}
\toprule
\multirow{2}{*}{Method} & \multirow{2}{*}{Input} & \multicolumn{2}{c}{\underline{2D Metrics}} &  \multicolumn{3}{c}{\underline{3D Metrics}}  \\
& & $\delta_{2D}<$1px & EPE & $\delta_{3D}<.05$ & $\delta_{3D}<0.10$ & EPE \\
\midrule
RAFT \cite{teed2020raft} & RGB & 79.4\% & 3.53 & - & - & - \\
RAFT (2D flow backprojected) & RGB-D & 78.8\% & 3.42 & 50.6\% & 55.7\% & 5.442 \\
RAFT (2D flow + depth change) & RGB-D & 75.2\% & 3.66 & 33.9\% & 47.2\% & 1.218 \\
RAFT (3D flow) & RGB-D & 73.6\% & 4.42 & 36.2\% & 55.4\% & 0.266 \\
Ours & RGB-D & \textbf{86.4}\% & \textbf{2.46} &  \textbf{87.8}\% & \textbf{91.5}\% & \textbf{0.062} \\
\midrule
\end{tabular}
}
\caption{Results on the FlyingThings3D dataset using the images from the FlowNet3D split. We evaluate on the full images (excluding pixels at infinity and extremely fast moving regions with flow $>250$px)}
\label{table:FlyingThingsResults}
\end{table*}

\section{Experiments}

We evaluate our approach on a variety of real and synthetic datasets. For all experiments we use the AdamW optimizer\cite{loshchilov2017decoupled} with weight decay set to $1\times10^{-5}$ and unroll 12 iterations of the update operator. All components of the network are trained from scratch, with the exception of the context encoder which uses ImageNet~\cite{deng2009imagenet} pretrained weights. 

Training RAFT-3D involves differentiating a computation graph which consists of both Euclidean tensors (e.g. network weights, feature activation) and Lie Groups elements (e.g. SE3 transformation field). We use the LieTorch library\cite{teed2021tangent} to perform backpropagation in the tangent space of manifold elements in the computation graph.

\begin{table}[h]
\centering
\resizebox{\columnwidth}{!}{
\begin{tabular}{llccc}
\toprule
Method & Input & $\delta_{3D}<.05$ & $\delta_{3D}<0.10$ & EPE$_{3D}$ \\
\midrule
FlowNet3D \cite{liu2019flownet3d} & XYZ & 25.4\% & 57.9\% & 0.169 \\
FlowNet3D++\cite{wang2020flownet3d++} & RGB-D & 30.3\% & 63.4\% & \underline{0.137} \\
FLOT\cite{puy20flot} & XYZ & \underline{34.3}\% & \underline{64.3}\% & 0.156 \\
\midrule
Ours & RGB-D & \textbf{83.7}\% & \textbf{89.2}\% & \textbf{0.064} \\
\midrule
\end{tabular}
}
\caption{3D scene flow results on the FlyingThings3D dataset using the split proposed by Liu et al \cite{liu2019flownet3d} where only non-occluded points with depth $<$35m are considered for evaluation. Our method outperforms existing point-based scene flow networks by a large margin.}
\label{table:FlyingThingsResults3D}
\end{table}

\begin{table*}[tb]
\centering
\resizebox{.85\textwidth}{!}{
\begin{tabular}{lccccccccccccc}
\toprule
&&\multicolumn{3}{c}{Disparity 1}&\multicolumn{3}{c}{Disparity 2}&\multicolumn{3}{c}{Optical Flow} &\multicolumn{3}{c}{Scene Flow}\\
Methods & Runtime &\emph{bg} &\emph{fg} &{all} &\emph{bg} &\emph{fg} &{all} &\emph{bg} &\emph{fg} &{all} &\emph{bg} &\emph{fg} &{all}\\
\hline
OSF \cite{menze2015object} &50 mins&4.54&12.03&5.79&5.45&19.41&7.77&5.62&18.92&7.83&7.01&26.34&10.23\\
SSF \cite{ren2017cascaded} &5 mins& 3.55& 8.75& 4.42& 4.94& 17.48& 7.02& 5.63& 14.71& 7.14& 7.18& 24.58& 10.07\\
Sense \cite{jiang2019sense} & 0.31s & 2.07 & 3.01 & 2.22 & 4.90 & 10.83 & 5.89 & 7.30 & 9.33 & 7.64 & 8.36 &	15.49 & 9.55 \\	
DTF Sense \cite{schuster2020deep} & 0.76 sec & 2.08 & 3.13 & 2.25 & 4.82 & 9.02 & 5.52 & 7.31 & 9.48& 7.67 & 8.21 & 14.08 & 9.18 \\	
PRSM* \cite{vogel20153d} & 5 mins &3.02&10.52&4.27&5.13&15.11&6.79&5.33&13.40&6.68&6.61&20.79&8.97\\
OpticalExp \cite{yang2020upgrading} & 2.0 sec & \textbf{1.48} & \textbf{3.46} & \textbf{1.81} & 3.39 & \textbf{8.54} & 4.25 & 5.83 & \textbf{8.66} & 6.30 & 7.06 & 13.44 & 8.12 \\
ISF \cite{behl2017bounding} & 10 mins&4.12&6.17&4.46&4.88&11.34&5.95&5.40&10.29&6.22&6.58&15.63&8.08\\
ACOSF \cite{Cong2020ICPR} & 5mins & 2.79 & 7.56 & 3.58 & 3.82 & 12.74 & 5.31 & 4.56 & 12.00 & 5.79 & 5.61 & 19.38 & 7.90 \\
DRISF\cite{ma2019deep} & 0.75 sec (2 GPUs) & 2.16 & 4.49 & 2.55 & 2.90 & 9.73 & 4.04 & 3.59 & 10.40 & 4.73 & 4.39 & 15.94 & 6.31 \\
\midrule
Ours & 2.0 sec & \textbf{1.48} & \textbf{3.46} & \textbf{1.81} & \textbf{2.51} & 9.46 & \textbf{3.67} & \textbf{3.39} & 8.79 & \textbf{4.29} & \textbf{4.27} & \textbf{13.27} & \textbf{5.77} \\
\midrule
\end{tabular}
}
\caption{Results of the top performing methods on the KITTI leaderboard. Ours ranks first on the leaderboard among all published methods.}
\label{tab:kittiresults}
\end{table*}

\subsection{FlyingThings3D}
The FlyingThings3D dataset was introduced as part of the synthetic Scene Flow datasets by Mayer et al. \cite{mayer2016large}. The dataset consists of ShapeNet \cite{chang2015shapenet} shapes with randomized translation and rotations placed in a scene populated with background objects. While the dataset is not naturalistic, it offers a challenging combination of camera and object motion, each of which span all 6 degrees of freedom.

We train our network for 200k iterations with a batch size of 4 and a crop size of [320, 720]. We perform spatial augmentation by random cropping and resizing and adjust intrinsics accordingly. We use an initial learning rate of .0001 and decay the learning rate linearly during training. 

We evaluate our network using 2D and 3D end-point-error (EPE). 2D EPE is defined as the euclidean distance between the ground truth optical flow and the predicted optical flow which can be obtained from the 3D transformation field using Eqn. \ref{eqn:mapping}. 3D EPE is the euclidean distance between the ground truth 3D scene flow and the predicted scene flow. We also report threshold metrics, which measure the portion of pixels which lie within a given threshold.

In Tab. \ref{table:FlyingThingsResults3D} we compare to point cloud based scene flow methods\cite{liu2019flownet3d,wang2020flownet3d++,puy20flot} using the split proposed in FlowNet3D\cite{liu2019flownet3d} containing roughly 2000 test examples sampled from the FlyingThings3D test set. In this evaluation setup, only non-occluded pixels with depth $<$35 meters are used for evaluation. Our method improves the 3D $\delta<0.05$ accuracy from 34.3\% to 83.7\%.

In Tab. \ref{table:FlyingThingsResults} we compare to RAFT\cite{teed2020raft} and several baselines we implement to extend RAFT to predict 3D motion. All RAFT baselines use the same network architecture as our approach, including the pretrained ResNet-50. All baselines are provided with inverse depth as input which is concatenate with the input images. We also experiment with directly provided depth as input, but found that inverse depth gives the best results.

RAFT (2D flow backprojected) uses the depth maps to backproject 2D motion into a 3D flow vector, but this only works for non-occluded pixels, which is the reason for the very large 3D EPE error. RAFT (2D flow + depth change) predicts 2D flow in addition to inverse depth change, which can be used to recover 3D flow fields. Finally, we also test a version of RAFT which predicts 3D motion fields directly; RAFT(3D flow). We find that our method outperforms all these baselines by a large margin, particularly on the 3D metrics. This is because our network operates directly on the SE3 motion field, which offers a more structured representation than flow fields and we produce analytically constrained updates which the other baselines lack.

In this experiment, we evaluate over all pixels (excluding \emph{extremely} fast moving objects with flow $>$250 pixels). Since we decompose the scene into rigidly moving components, our method can estimate the motion of occluded regions as well. We provide qualitative results in Fig.~\ref{fig:examples}. These examples show that our network can segment the scene into rigidly moving regions, producing piecewise constant SE3 motion fields, even though no supervision is used on the embeddings.

\subsection{KITTI}
Using our model trained on FlyingThings3D, we finetune on KITTI for an additional 50k iterations with an initial learning rate of $5\times10^{-5}$. We use a crop size of [288, 960] and perform spatial and photometric augmentation. To estimate disparity, we use GA-Net\cite{zhang2019ga}, which provides the input depth maps for our method.

\begin{table*}
\centering
\resizebox{.68\textwidth}{!}{
\begin{tabular}{clccccc}
\toprule
\multirow{2}{*}{Experiment} & \multirow{2}{*}{Configuration} &  \multicolumn{2}{c}{\underline{2D Metrics}} &  \multicolumn{3}{c}{\underline{3D Metrics}}  \\
& & $\delta_{2D}<$1px & EPE & $\delta_{3D}<.05$ & $\delta_{3D}<0.10$ & EPE \\
\midrule
\multirow{5}{*}{Iterations}
& 1 & 62.1 & 6.05 & 56.0 & 65.9 & 0.212 \\
& 3 & 82.8 & 2.95 & 80.5 & 85.7 & 0.098 \\
& 8 & 85.5 & 2.47 & 86.4 & 90.5 & 0.062 \\
& \underline{16} & \textbf{85.8} & \textbf{2.43} & \textbf{87.1} & \textbf{91.0} & \textbf{0.059} \\
& 32 & 85.7 & 2.50 & 87.0 & 90.9 & 0.061 \\
\midrule
\multirow{4}{*}{Neighborhood Radius (px) }
& 8 & 73.2 & 4.01 & 38.7 & 59.0 & 0.192 \\
& 64 & 83.8 & 2.52 & 78.1 & 86.6 & 0.078 \\
& \underline{256} & \textbf{85.8} & \textbf{2.43} & \textbf{87.1} & \textbf{91.0} & \textbf{0.059} \\
& Full Image &  83.3 & 2.91 & 83.2 & 88.1 & 0.078 \\
\midrule
\multirow{2}{*}{Revision Factors}
& Flow & \textbf{86.1} & \textbf{2.29} & 84.6 & 88.7 & 0.081 \\
& \underline{Flow + Inv. Depth} & 85.8 & 2.43 & \textbf{87.1} & \textbf{91.0} & \textbf{0.059} \\
\midrule
\multirow{2}{*}{bi-Laplacian Smoothing}
& No & 85.8 & \textbf{2.43} & 87.1 & 91.0 & \textbf{0.059} \\
& \underline{Yes} &  \textbf{86.3} & 2.45 & \textbf{87.8} & \textbf{91.5} & 0.062\\
\bottomrule
\end{tabular}
}
\caption{Ablation experiments, details of the individual experiments are provided in \ref{sec:ablations}}
\label{table:ablations}
\end{table*}

We submit our method to the KITTI leaderboard and report results from our method and other top performing methods in Tab. \ref{tab:kittiresults}. Our approach outperforms all published methods. DRISP \cite{ma2019deep} is the next best performing approach, and combines PSMNet\cite{chang2018pyramid}, PWC-Net\cite{pwcnet}, and Mask-RCNN\cite{he2017mask}. Mask-RCNN is pretrained on Cityscapes and fine-tuned on KITTI using bounding box and instance mask supervision. Our network outperforms DRISP despite only training on FlyingThings3D and KITTI, and uses no instance supervision.

\subsection{Ablations}
\label{sec:ablations}

We ablate various components of our model on the FlyingThings dataset and report results in Tab. \ref{table:ablations}. For all ablations, we use our network without bi-Laplacian optimization as the baseline architecture.

\vspace{1mm} \noindent \textbf{Iterations:} We evaluate the performance of our model as function of the number of application of the update operator. We find that more iterations gives better performance up to about 16, after which we observe a slight degradation.

\vspace{1mm} \noindent \textbf{Neighborhood Radius:} The Dense-SE3 layer defines an objective which such at all pairs of pixels within a specific radius $r$ contribute to the objective. Here, we train networks where $r$ is set to $\{8, 64, 256, \infty\}$. In the last case, all pairs of pixels in the image contribute to the objective. We find that  $256$ gives the better performance than smaller radii; however, using the full image gives worse performance. This is likely due to the fact that most rigid objects will be less than 512 pixels in diameter, and imposing a restriction on the radius is a useful prior.

\vspace{1mm} \noindent \textbf{Revision Factors:} The update operator produces a set of revisions which are used as input to the Dense-SE3 layer. Here we experiment with different revisions. In \emph{Flow} we only use the optical flow revisions $\mathbf{r}_x$ and $\mathbf{r}_y$. In \emph{flow + inv. depth} we include inverse depth revisions. We find that including inverse depth revisions leads to better performance on 3D metrics because it leverages depth consistency.

\begin{figure}
    \centering
    \small with bi-Laplacian \hspace{10mm} without bi-Laplacian
    \includegraphics[width=.9\columnwidth]{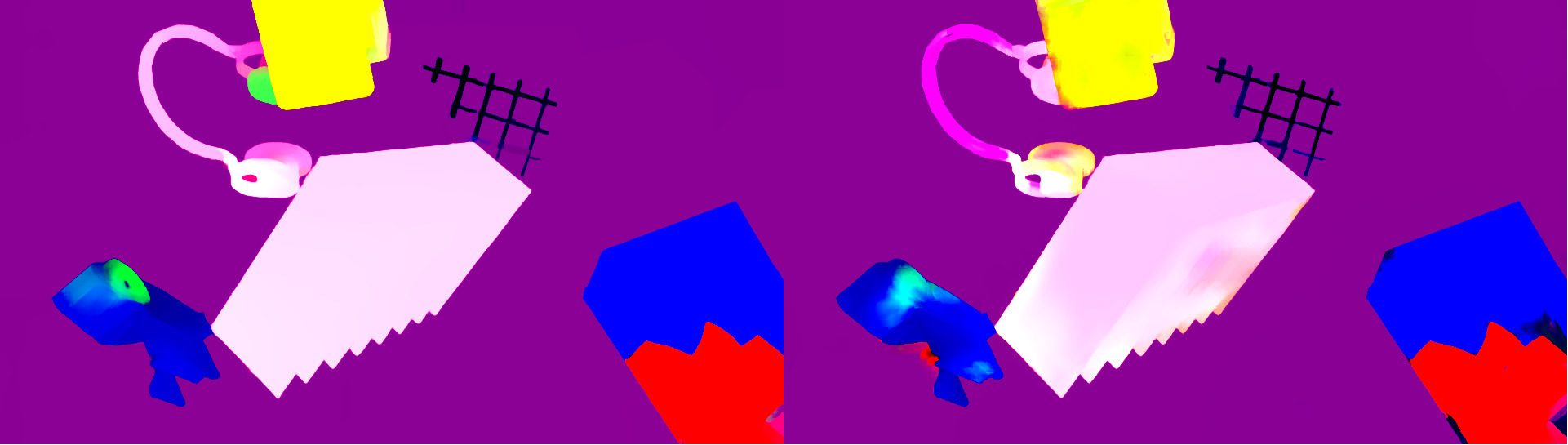}
    \caption{Impact of bi-Laplacian optimization layer on motion fields. This layer improves the ability of the network to aggregate embedding vectors within motion boundaries.}
    \label{fig:gbap}
    \vspace{-4mm}
\end{figure}

\vspace{1mm} \noindent \textbf{bi-Laplacian Optimization:} Here we test the impact bi-Laplacian optimization layer. Our pooling layer improves the accuracy of the threshold metrics improving 1px accuracy from 85.8 to 86.3, and 3D accuracy from 87.1 to 87.8 and gives comparable average EPE. In Fig. \ref{fig:gbap} we see that the pooling layer produces qualitatively better results, particularly over large objects.

\vspace{1mm}
\noindent \textbf{Parameter Count and Timing: } RAFT-3D has 45M trainable parameters. The ResNet50 backbone has 40M parameters, while the feature extractor and update operator make up the remaining 5M parameters. 

We provide a breakdown of the inference time in Tab. \ref{table:timing}. Timing results are computed on 540x960 images with a GTX 1080Ti GPU using 16 updates. Inference on 540x960 images requires 1.6G of GPU memory, which is mainly required to store the 4D correlation volume.

\begin{table}[h]
\centering
\resizebox{.6\columnwidth}{!}{
\begin{tabular}{ll}
\toprule
Component & Time (ms) \\
\midrule
Feature Extraction & 52ms\\
Cost Volume & 4ms \\
Update Operator (GRU) & 208ms (13ms/iter) \\
Gauss Newton Iteration & 120ms (7.5ms/iter) \\
SE3 Upsampling & 2ms \\
\midrule
Total & 386ms \\
\midrule
\end{tabular}
}
\caption{Forward pass timing for different components.}
\label{table:timing}
\end{table}

\vspace{-5mm}
\section{Conclusion}
We have introduced RAFT-3D, an end-to-end network for scene flow. RAFT-3D uses rigid-motion embeddings, which represent a soft grouping of pixels into rigidly moving objects. We demonstrate that these embeddings can be used to solve for dense and accurate 3D motion fields.

\vspace{2mm} \noindent \textbf{Acknowledgements:} This research is partially supported by the National Science Foundation under Grant IIS-1942981.

{\small
\bibliographystyle{ieee_fullname}
\bibliography{egbib}
}








\appendix

\begin{center}
    \Large \textbf{RAFT-3D: Appendix}
\end{center}

\begin{figure}[h]
    \centering
    \includegraphics[width=\linewidth]{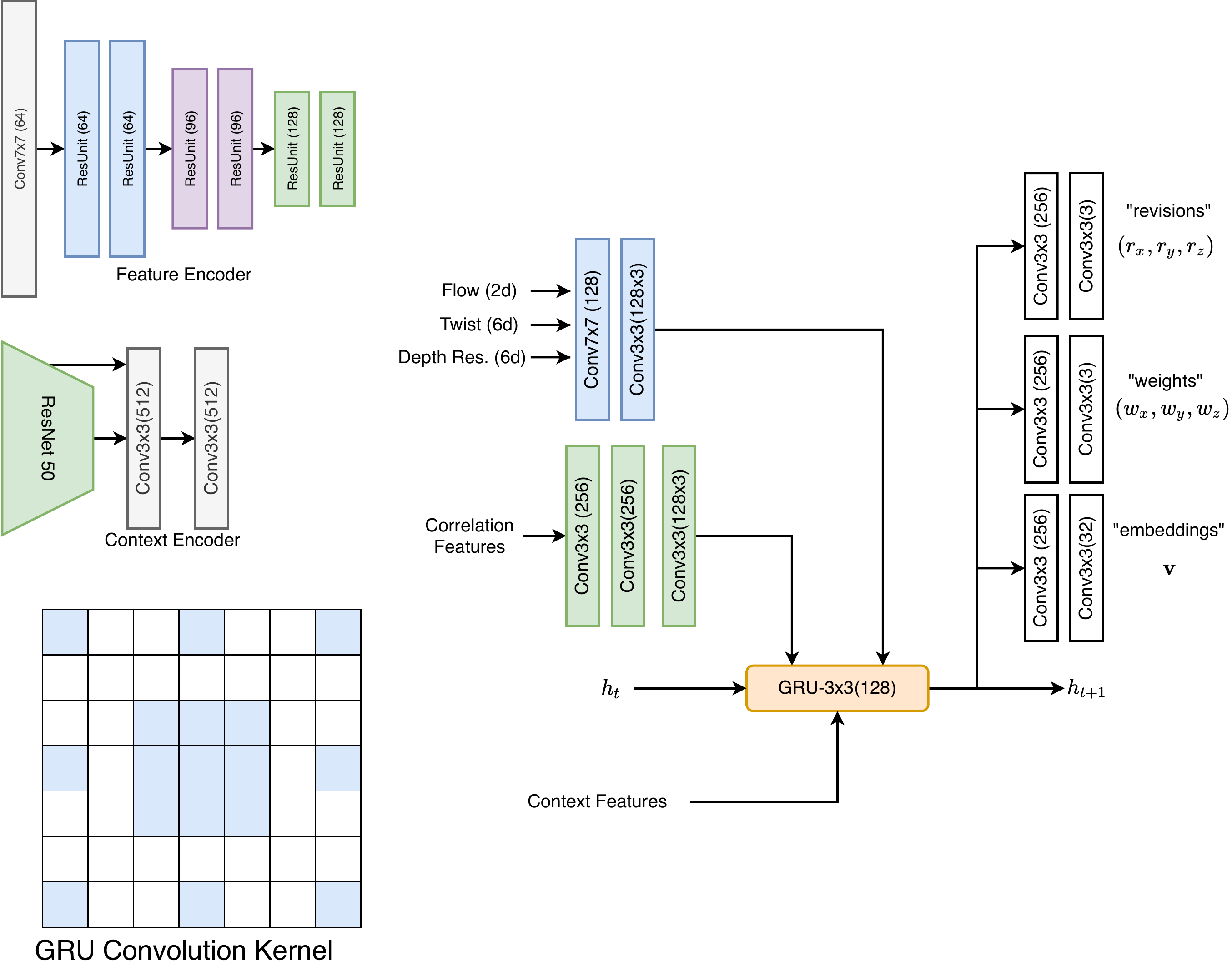}
    \caption{Network architecture. The components include (1) a feature encoder (2) a context encoder with a ResNet50 backbone, and (3) and GRU-based updated operator. The GRU uses a dilated convolution pattern as shown. In contrast to RAFT\cite{teed2020raft} where features are concatenated before being passed to the GRU, we perform elementwise addition of the context features, correlation features, and motion features.}
    \label{fig:architecture}
\end{figure}

\section{Network Architecture}

Details of the network architecture, including feature encoders and the GRU-based update operator are shown in Figure \ref{fig:architecture}.

\section{bi-Laplacian Optimization Layer Gradients}

This layer minimizes an objective function in the form
\begin{equation}
    ||D_x \mathbf{u}||_{\mathbf{w}_x}^2 + ||D_x \mathbf{u}||_{\mathbf{w}_y}^2 + ||\mathbf{u} - \mathbf{v}||^2
\end{equation}
where $D_x$ and $D_y$ are linear finite difference operators, and $\mathbf{v}$ is the flattened feature map. 

First consider the case of single channel, $\mathbf{v}\in \mathbb{R}^{HW}$. Let $W_x = \diag(\mathbf{w}_x), W_y = \diag(\mathbf{w}_y) \in \mathbb{R}^{HW\times HW}$. We can solve for $\mathbf{u}^*$ 
\begin{equation}
    (\mathbf{I} + D_x^T W_x D_x^T + D_y^T W_y D_y^T)\mathbf{u}^* = \mathbf{v}
    \label{eqn:pooling}
\end{equation}
We perform sparse Cholesky factorization and backsubstition to solve for $\mathbf{u}^*$ using the Cholmod library\cite{chen2008algorithm}.

\vspace{1mm} \noindent \textbf{Gradients:} In the backward pass, given the gradient $\frac{\partial L}{\partial\mathbf{u}^*}$, we need to find the gradients with respect to the boundary weights $\frac{\partial L}{\partial\mathbf{w}_x}$ and $\frac{\partial L}{\partial\mathbf{w}_y}$.

Given the linear system $\mathbf{H} \mathbf{u} = \mathbf{v}$, the gradients with respect to $\mathbf{H}$ and $\mathbf{v}$ can be found by solving the system in the backward direction \cite{amos2017optnet}
\begin{align}
    \frac{\partial L}{\partial \mathbf{v}} &= \mathbf{H}^{-T} \frac{\partial L}{\partial\mathbf{u}^*} \\
    \frac{\partial L}{\partial \mathbf{H}} &= \mathbf{u}^* \mathbf{d}_v^T \\
    \mathbf{d}_v &= \mathbf{H}^{-T} \frac{\partial L}{\partial\mathbf{u}^*}
\end{align}
Here the column vector $\mathbf{d}_v$ is defined for notational convenience. Since $\mathbf{H}$ is positive definite,  $\mathbf{H}^{-T}=\mathbf{H}^{-1}$ so we can reuse the factorization from the forward pass.

To compute the gradients with respect to $\mathbf{w}_x$ and $\mathbf{w}_x$
\begin{align}
    \frac{\partial L}{\partial \mathbf{w}_x} = \diag\left( \frac{\partial L}{\partial \mathbf{H}} \frac{\partial \mathbf{H}}{\partial W_x} \right) \\
    = \diag\left((D_x \mathbf{u}^*)(D_x \mathbf{d}_v)^T \right)
\end{align}
giving
\begin{equation}
    \frac{\partial L}{\partial \mathbf{w}_x} = (D_x \mathbf{u}^*) \odot (D_x \mathbf{d}_v)
\end{equation}
where $\odot$ is elementwise multiplication. Similarly
\begin{equation}
    \frac{\partial L}{\partial \mathbf{w}_y} = (D_y \mathbf{u}^*) \odot (D_y \mathbf{d}_v)
\end{equation}

\vspace{1mm} \noindent \textbf{Multiple Channels:} We can easily extend Eqn. \ref{eqn:pooling} to work with multiple channels. Since the matrix $\mathbf{H}$ does not depend on $\mathbf{v}$, it only needs to be factored once. We can solve Eqn. \ref{eqn:pooling} for all channels by reusing the factorization, treating $\mathbf{v}$ as a $HW \times C$ matrix. The gradient formulas can also be updated by summing the gradients over the channel dimensions.



\end{document}